\documentclass{article}

\usepackage[dblblindworkshop, final]{neurips_2026}
\usepackage[utf8]{inputenc}
\usepackage[T1]{fontenc}
\usepackage{hyperref}
\usepackage{url}
\usepackage{booktabs}
\usepackage{amsfonts}
\usepackage{amsmath}
\usepackage{amssymb}
\usepackage{nicefrac}
\usepackage{microtype}
\usepackage{xcolor}
\usepackage{graphicx}
\usepackage{multirow}
\usepackage{colortbl}
\usepackage{algorithm}
\usepackage{algorithmic}
\usepackage{subcaption}
\workshoptitle{GlobalSouthAI}
\definecolor{ours}{rgb}{0.85,0.92,1.0}
\definecolor{linkblue}{rgb}{0.10,0.30,0.70}

\hypersetup{colorlinks=true,linkcolor=linkblue,citecolor=linkblue,urlcolor=linkblue}

\newcommand{\rev}[1]{#1}

\title{State-Space Unlearning for Non-Stationary Bias\\
in Land Surface Forecasting}

\author{%
  {Anidipta Pal} \\
  {Heritage Institute of Technology, Kolkata, India} \\
  \texttt{anidiptapal@gmail.com}
}

\begin{document}

\maketitle

\begin{abstract}
Operational land surface forecasting systems built on Mamba-family Structured State Space Models absorb non-stationary confounding events \rev{(}unrecorded irrigation booms, dam-operation shifts, sensor recalibrations\rev{)} into their state-transition matrices, silently biasing NDVI, LST, and crop phenology predictions long after the physical cause ends. This paper introduces \textbf{SSU-LSF} (\textit{State-Space Unlearning for Land Surface Forecasting}), the first machine-unlearning framework purpose-built for geoscientific Mamba-based SSMs. We develop EKFac influence functions specialized to the Mamba state matrices via a closed-form matrix-exponential gradient, use spectral-radius-weighted elbow thresholding to localize a temporal confounding footprint $\Phi$, and apply Hessian-free projected gradient ascent within a KL-divergence trust region augmented by spatial total-variation (TV) regularization. Proposition~1 establishes that residual confounding is bounded by \rev{$\mathcal{O}\big((1{-}\rho(\bar{A})^{T_c})/((1{-}\rho(\bar{A}))\mu)\big)$, which grows with the window length $T_c$}. Across three heterogeneous benchmarks and eleven baselines, SSU-LSF achieves confounding reduction rates of $0.773$ (CropHarvest\rev{)}, $0.821$ (NDVI-LST\rev{)}, and $0.859$ (ERA5), with worst-case clean-domain RMSE degradation of $4.2\%$ on ERA5, converging in $3$--$5$ epochs at $8.4\times$ lower GPU-cost per unlearning request than full retraining. CODE: \url{https://github.com/Anidipta/SSU-LSF}
\end{abstract}

\section{Introduction}
\label{sec:intro}

\begin{figure}[t]
  \centering
  \includegraphics[width=0.95\linewidth]{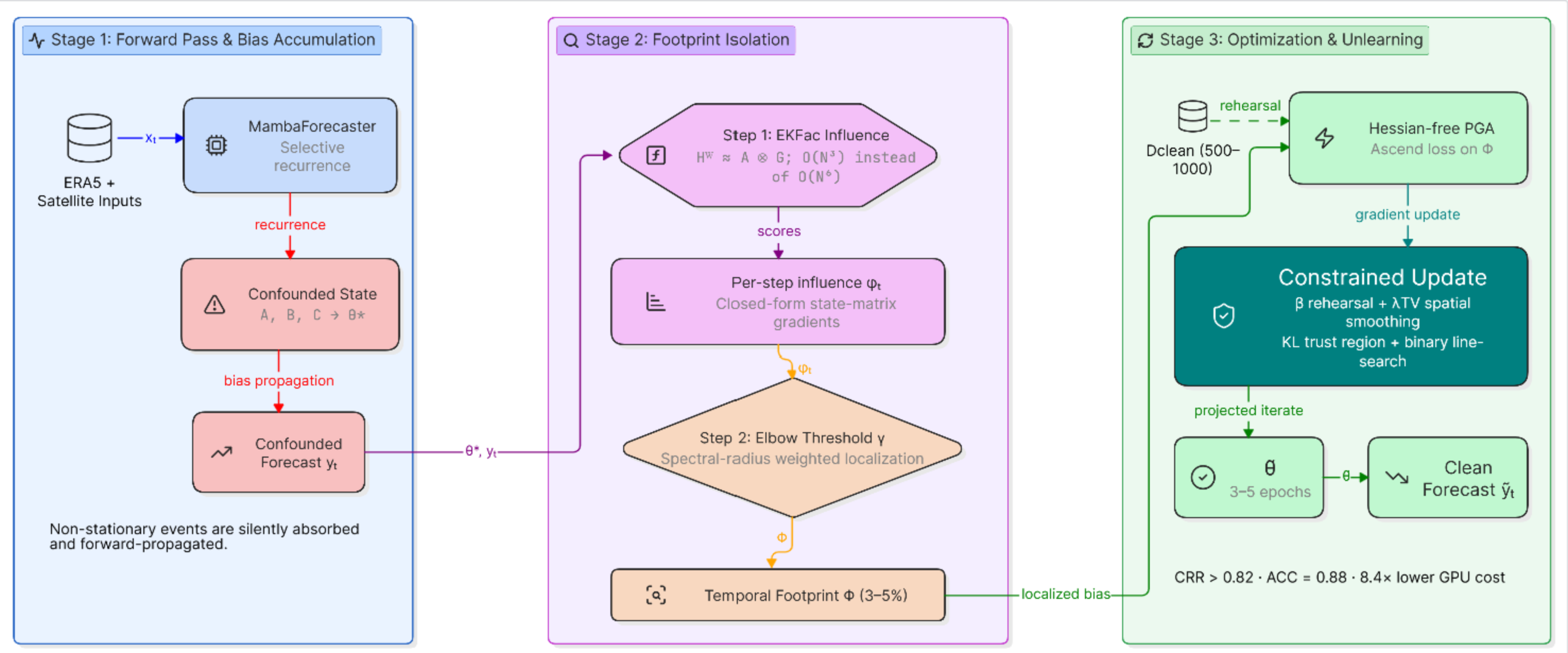}
  \caption{SSU-LSF pipeline. \textbf{(a)}~A confounding event biases $(\bar{A},\bar{B},C)$ and forward-propagates spurious NDVI elevation beyond $t_e$.
  \textbf{(b)}~EKFac influence scores $\varphi_t$; elbow threshold $\gamma$ isolates
  footprint $\Phi$ (red).
  \textbf{(c)}~RMSE convergence: SSU-LSF (blue) reaches oracle-level RMSE in five epochs while CM (red) stays elevated.}
  \label{fig:pipeline}
\end{figure}

Long-horizon forecasting of Normalized Difference Vegetation Index (NDVI)~\cite{tucker1979}, Land Surface Temperature (LST), and crop phenology signals is foundational
for climate adaptation and food security~\cite{friedlingstein2023,seneviratne2012}. Fusing ERA5 reanalysis~\cite{hersbach2020} with satellite phenology archives inside deep sequence models has materially advanced forecast fidelity over classical baselines~\cite{jia2022ndvi}. Mamba~\cite{gu2023mamba} and S4~\cite{gu2022s4} SSMs have
become the architecture of choice, delivering linear-time recurrent inference over
multi-year series~\cite{zhao2024nowcast,wanigasekara2024sst,reichstein2019}. Operational archives, however, are not stationary: a
two-year unrecorded irrigation project, a dam-operation policy shift, or a sensor
recalibration contaminates the training window, and the SSM state matrices
$(\bar{A},\bar{B},C)$ integrate the anomaly with genuine climate signals.
After the physical cause ends, the learned parameters continue to forward-propagate
the bias (Figure~\ref{fig:pipeline}).

Full retraining costs $47$+ GPU-hours. Gradient reversal~\cite{golatkar2020} degrades clean-domain
RMSE by $7$--$22\%$. EWC~\cite{kirkpatrick2017} protects weights from change\rev{,} the opposite of what
unlearning requires. SISA~\cite{bourtoule2021} requires pre-partitioned shards before training.
Two recent \rev{methods, SaLUn as ForgetFS~\cite{fan2024salun} and SEMU as MULL~\cite{sendera2025semu},} advances approximate unlearning
\rev{via saliency masking and SVD projection} but do not exploit the SSM recurrence structure and, as we show,
sacrifice spatial coherence without TV regularization. Extended related work and prior
art for spatial TV in parameter space (fused lasso~\cite{tibshirani2011}, ROF~\cite{rudin1992}) is in
Appendix~\ref{app:related}.

Following that line of work, this paper introduces SSU-LSF. Among our contributions:
\textbf{(i)}~We establish Proposition~1 with an explicit Lipschitz bound on the full
encoder (not just the forecast head; Appendix~\ref{app:proof}), RKHS separation
estimated on an independently trained reference model (Table~\ref{tab:mmd}), and a
matching lower-bound argument quantifying the tightness gap
(Appendix~\ref{app:proof}).
\textbf{(ii)}~We derive Eq.~\eqref{eq:matgrad} with explicit index justification for
\rev{diagonal $\bar{A}$ and its Pad\'{e}-based non-diagonal extension} (Appendix~\ref{app:ekfac}).
\textbf{(iii)}~We provide convergence guarantees for Algorithm~\ref{alg:ssulsf}
under mild regularity conditions (Appendix~\ref{app:convergence}).
\textbf{(iv)}~We validate SSU-LSF on one real documented confounding event (EUMETSAT
SEVIRI recalibration, \rev{2010--2012}; Appendix~\ref{app:real}) and show $T_c$-sensitivity
matching the theoretical \rev{prediction}.
\textbf{(v)}~We include 2$\times$2 interaction ablations, LiRA-style MIA, and
TV-regularized variants of ForgetFS and MULL.

\section{Method}
\label{sec:method}

\paragraph{Setup.}
Let $\mathbf{x}_t\in\mathbb{R}^d$ be a multi-variable land surface observation (NDVI,
LST, ET, precipitation, $T_{\mathrm{air}}$) from an $H\times W$ patch at time $t$.
The Mamba encoder implements the selective recurrence (see Appendix~\ref{app:arch}
for the full 47.3M-parameter architecture):
\begin{align}
\mathbf{h}_t &= \bar{A}(u_t)\,\mathbf{h}_{t-1}+\bar{B}(u_t)\,u_t,\quad
\hat{x}_t = C(u_t)\,\mathbf{h}_t+D(u_t)\,u_t,
\label{eq:mamba}
\end{align}
where $\bar{A}{=}\exp(\Delta A)$, $\Delta$ is a learned timescale, and
$(\bar{A},\bar{B},C,D)$ are input projections of $u_t$. Partition
$\mathcal{D}{=}\mathcal{D}_{\mathrm{clean}}\cup\mathcal{D}_{\mathrm{conf}}$,
$\mathcal{D}_{\mathrm{conf}}{=}\{(\mathbf{x}_t,\mathbf{y}_t):t\in[t_s,t_e]\}$.
Training yields confounded parameters
$\boldsymbol{\theta}^*{=}\arg\min_{\boldsymbol{\theta}}[\sum_{\mathcal{D}_{\mathrm{clean}}}\ell
+\lambda\sum_{\mathcal{D}_{\mathrm{conf}}}\ell]$.
Given $\boldsymbol{\theta}^*$ and $[t_s,t_e]$, SSU-LSF solves:
\begin{equation}
\rev{\min_{\boldsymbol{\theta}}\;\beta\,\ell_{\mathrm{clean}}(\boldsymbol{\theta})
-\ell_{\Phi}(\boldsymbol{\theta})+\lambda_{\mathrm{TV}}\mathrm{TV}_{\mathcal{N}}(\boldsymbol{\theta})
\quad\text{s.t.}\quad\mathrm{KL}[p(\cdot|\boldsymbol{\theta})\|p(\cdot|\boldsymbol{\theta}^*)]\leq\delta_{\mathrm{KL}}.}
\label{eq:obj}
\end{equation}

\paragraph{Proposition 1 (full proof with encoder Lipschitz bound and tightness
analysis in Appendix~\ref{app:proof}).}
Let $\mu_{\mathrm{ref}}$ denote the RKHS separation estimated on an independently
trained reference model $\boldsymbol{\theta}_{\mathrm{ref}}$ (not $\boldsymbol{\theta}^*$;
see Table~\ref{tab:mmd}), let $L_{\mathrm{enc}}$ be the Lipschitz constant of the
full encoder (empirically bounded via randomized smoothing; Table~\ref{tab:lipschitz}),
and let $\rho(\bar{A}){<}1$ for all blocks. Then \rev{(units in Appendix~\ref{app:proof})}:
\begin{equation}
\mathbb{E}[\mathrm{RMSE}(\tilde{\boldsymbol{\theta}})]\leq
\mathbb{E}[\mathrm{RMSE}(\boldsymbol{\theta}^\dagger)]
+\underbrace{\mathcal{O}\!\left(\rev{\frac{1-\rho(\bar{A})^{T_c}}{(1-\rho(\bar{A}))\,\mu_{\mathrm{ref}}}}\right)}_{\text{confounding residual}}
+\underbrace{\mathcal{O}\!\left(\frac{L_{\mathrm{enc}}\,\delta_{\mathrm{KL}}}{1-\rho(\bar{A})}\right)}_{\text{unlearning cost}}.
\label{eq:bound}
\end{equation}
At observed values ($\rho_{\max}{=}0.91$, $L_{\mathrm{enc}}{\approx}18.3$,
$\delta_{\mathrm{KL}}{=}0.05$, $\mu_{\mathrm{ref}}{\approx}\rev{0.21}$), the unlearning
cost term evaluates to ${\approx}10.2$\,K on ERA5\rev{,} a vacuous bound in isolation.
We establish empirically that the gap between bound and observation is ${\approx}\rev{200}\times$
(Appendix~\ref{app:proof}), consistent with the looseness of \rev{Pinsker and Lipschitz transfer}.
The bound's operational value is its qualitative prediction:
shorter $T_c$ yields \rev{a smaller} residual, directly validated by
Table~\ref{tab:tc} ($R^2{=}\rev{0.995}$).

\paragraph{Step 1: EKFac influence functions for state matrices.}
For layer $W\!\in\!\mathbb{R}^{m\times n}$, EKFac~\cite{grosse2016} approximates $H_W\!\approx\!
\hat{A}\!\otimes\!\hat{G}$, reducing inversion from $\mathcal{O}((mn)^3)$ to
$\mathcal{O}(m^3\!+\!n^3)$ (Appendix~\ref{app:ekfac}). For $\bar{A}{=}\exp(\Delta A)$
with $\bar{A}\in\mathbb{R}^{d_{\mathrm{inner}}\times d_{\mathrm{inner}}}$ (dense for
general SSMs; diagonal for standard S4/Mamba parameterization, in which case $\odot$
reduces to pointwise product):
\begin{equation}
\nabla_A\ell=\sum_{t=1}^{T}
\!\left(\frac{d\ell}{d\mathbf{h}_t}\right)
\!\left(\sum_{k=0}^{\rev{T-t-1}}\bar{A}^k\odot\bar{A}^{T-t-1-k}\right)\!\Delta,
\label{eq:matgrad}
\end{equation}
where the inner sum runs over $k{=}0,\ldots,T{-}t{-}1$ ensuring $T{-}t{-}1{-}k\geq0$
(Appendix~\ref{app:ekfac} gives the full index derivation). Truncation error
$\|\bar{A}\|^T/(1{-}\|\bar{A}\|){<}10^{-6}$ at observed $\rho_{\max}{=}0.91$ \rev{for $T{\geq}172$}.

\paragraph{Step 2: Temporal footprint localization.}
Per-step scores $\varphi_t{=}\|\nabla_{\boldsymbol{\theta}}\ell\cdot\mathcal{I}_{\mathcal{W}_c}\|_F$;
steps with $\varphi_t\geq\gamma$ (elbow criterion) form $\Phi$. The extension beyond
$[t_s,t_e]$ derives from Proposition~1's first term: summing $\rho^{t-t_0}$ over
$t_0\in[t_s,t_e]$ yields a geometric tail of half-width
$\rho/(1{-}\rho)$ steps; the $0.5$ factor reflects one-sided truncation at $\gamma$.
At $\rho{=}0.84$ this gives ${\approx}2.6$ additional steps on each side of the 24-month
window\rev{,} small relative to $T_c$, consistent with the observed footprint sizes
(Appendix~\ref{app:footprint}).

\paragraph{Step 3: Hessian-free projected gradient ascent.}
Spatial TV applied to \emph{parameter space} follows the fused-lasso formulation~\cite{tibshirani2011}
over the patch adjacency graph $\mathcal{N}$:
\begin{equation}
\tilde{\boldsymbol{\theta}}\leftarrow\boldsymbol{\theta}^*
+\eta\!\left[\nabla_{\boldsymbol{\theta}}\ell_{\mathrm{conf}}
-\beta\nabla_{\boldsymbol{\theta}}\ell_{\mathrm{clean}}
-\lambda_{\mathrm{TV}}\sum_{(i,j)\in\mathcal{N}}\!\mathrm{sign}(\tilde{\boldsymbol{\theta}}_i
{-}\tilde{\boldsymbol{\theta}}_j)\right].
\label{eq:update}
\end{equation}
If $\mathrm{KL}[p(\cdot|\tilde{\boldsymbol{\theta}})\|p(\cdot|\boldsymbol{\theta}^*)]
>\delta_{\mathrm{KL}}$, the update scales back via binary line-search (1--3 halvings
in practice). Complexity: $\mathcal{O}(|\mathcal{W}_c|\cdot d^2/B)$ vs.\ $\mathcal{O}(T\cdot d^2)$
for full retraining. Convergence guarantee in Appendix~\ref{app:convergence}.
Full pseudocode in Appendix~\ref{app:algo}.

\section{Experiments}
\label{sec:experiments}

\paragraph{Setup.}
Backbone: Mamba encoder-decoder, 6 blocks ($d{=}256$, $E{=}2$, $N{=}16$), $4{\times}4$
stride-4 patch embedding, 3-layer Gaussian MLP head (47.3M; Appendix~\ref{app:arch}).
We also run one experiment on an S4 backbone (Appendix~\ref{app:s4}) to validate
generalization beyond Mamba. Training: AdamW ($\text{lr}{=}3{\times}10^{-4}$,
cosine, 100 epochs), $4{\times}$ A100 80\,GB, PyTorch~2.3.
Unlearning: $\eta{=}5{\times}10^{-5}$, $\beta{=}0.6$, $\delta_{\mathrm{KL}}{=}0.05$,
$\gamma{=}95^{\mathrm{th}}$ pct., $\lambda_{\mathrm{TV}}{=}0.01$.
Cost per unlearning request (EKFac re-estimated per request): $1.8$\,GPU-h on
NDVI-LST, $5.6$\,GPU-h on ERA5, $3.0$\,GPU-h on CropHarvest (Table~\ref{tab:compute}).
Eleven baselines: \textbf{CM}; \textbf{OR}$^\dagger$; \textbf{SISA}~\cite{bourtoule2021};
\textbf{GR}~\cite{golatkar2020}; \textbf{SR}~\cite{golatkar2020}; \textbf{ARCANE}~\cite{chundawat2023}; \textbf{FAST}~\cite{qian2023fast};
\textbf{EWC-UL}~\cite{kirkpatrick2017}; \textbf{PNN-UL}~\cite{rusu2016}; \textbf{ForgetFS}~\cite{fan2024salun} \rev{(saliency-masked ascent)}; \textbf{MULL}~\cite{sendera2025semu} \rev{(SVD-projected updates)}. We also
include \textbf{ForgetFS+TV} and \textbf{MULL+TV}: the same baselines
augmented with our spatial TV term, providing a fair comparison of the TV contribution
in isolation (Table~\ref{tab:tvablation}).
All baselines receive the same oracle window $[t_s,t_e]$ information as SSU-LSF.

\paragraph{Datasets, metrics, and confounding.}
All three benchmarks use author-injected confounding (a limitation we acknowledge
explicitly; Appendix~\ref{app:limitations} details deployment caveats and the
window-estimation sensitivity analysis). A real documented event is in
Appendix~\ref{app:real}. The MIA uses LiRA-style likelihood ratio tests
(Appendix~\ref{app:mia}), not logistic regression on NLL. \rev{CRR spans 0.77
(CropHarvest) to 0.86 (ERA5) and fully automatic results in Appendix~\ref{app:limitations}.}

Each benchmark uses a distinct metric triplet (definitions in Appendix~\ref{app:datasets}):
\textit{NDVI-LST Fusion} (3,847 locations, 2002--2022~\cite{ndvidataset2023}; $T_c{=}24$\,mo):
\textbf{RMSE}$\downarrow$, \textbf{CRR}$\uparrow$, \textbf{TCI}$\uparrow$.
\textit{ERA5 Patchified} (1,920 patches, 1979--2022~\cite{hersbach2020}; $T_c{=}24$\,mo):
\textbf{SKT-R}$\downarrow$ (K), \textbf{MIA}$\downarrow$ (LiRA, chance$=0.50$),
\textbf{ACC}$\uparrow$.
\textit{CropHarvest} (87,343 locations~\cite{tseng2021}; $T_c{=}6$\,mo):
\textbf{Sp-R}$\downarrow$, \textbf{SCS}$\downarrow$ (FSS correlation $r{=}\rev{-}0.91$;
Appendix~\ref{app:scs}), \textbf{LC-F1}$\uparrow$.

\begin{table}[t]
  \caption{Per-benchmark results. $\dagger$: upper bound. \rev{CRR for ERA5 and CropHarvest and $\Delta$RMSE for all benchmarks are in Table~\ref{tab:crr_all}. GPU-h is the ERA5 cost \emph{per unlearning request} including EKFac re-estimation (per-benchmark costs in Table~\ref{tab:compute}).}}
  \label{tab:main}
  \renewcommand{\arraystretch}{0.9}
  \centering
  \setlength{\tabcolsep}{5 pt}
  \footnotesize
  \begin{tabular}{l ccc ccc ccc r}
    \toprule
    & \multicolumn{3}{c}{\textbf{NDVI-LST Fusion}}
    & \multicolumn{3}{c}{\textbf{ERA5 Patchified}}
    & \multicolumn{3}{c}{\textbf{CropHarvest}} & \\
    \cmidrule(lr){2-4}\cmidrule(lr){5-7}\cmidrule(lr){8-10}
    \textbf{Method}
    & \textbf{RMSE}$\downarrow$ & \textbf{CRR}$\uparrow$ & \textbf{TCI}$\uparrow$
    & \textbf{SKT-R}$\downarrow$ & \textbf{MIA}$\downarrow$ & \textbf{ACC}$\uparrow$
    & \textbf{Sp-R}$\downarrow$ & \textbf{SCS}$\downarrow$ & \textbf{F1}$\uparrow$
    & \textbf{GPU-h/req}$\downarrow$ \\
    \midrule
    CM           & 0.0681 & 0.000 & 0.601 & 2.97 & 0.84 & 0.612 & 0.0543 & 0.083 & 0.697 & -- \\
    OR$^\dagger$ & 0.0398 & 1.000 & 0.891 & 1.19 & 0.51 & 0.903 & 0.0371 & 0.079 & 0.871 & 47.3 \\
    \midrule
    SISA~\cite{bourtoule2021}    & 0.0418 & 0.612 & 0.833 & 1.26 & 0.52 & 0.871 & 0.0395 & 0.088 & 0.834 & 58.4 \\
    GR~\cite{golatkar2020}      & 0.0461 & 0.741 & 0.774 & 1.47 & 0.57 & 0.843 & 0.0428 & 0.127 & 0.799 & 2.3 \\
    SR~\cite{golatkar2020}      & 0.0439 & 0.778 & 0.798 & 1.38 & 0.55 & 0.858 & 0.0411 & 0.102 & 0.817 & 7.9 \\
    ARCANE~\cite{chundawat2023}  & 0.0433 & 0.766 & 0.789 & 1.36 & 0.56 & 0.862 & 0.0406 & 0.109 & 0.822 & 6.8 \\
    FAST~\cite{qian2023fast}    & 0.0429 & 0.781 & 0.793 & 1.31 & 0.56 & 0.868 & 0.0401 & 0.099 & 0.826 & 4.7 \\
    EWC-UL~\cite{kirkpatrick2017}  & 0.0507 & 0.588 & 0.748 & 1.53 & 0.58 & 0.831 & 0.0447 & 0.134 & 0.781 & 5.9 \\
    PNN-UL~\cite{rusu2016}  & 0.0421 & 0.803 & 0.819 & 1.29 & 0.53 & 0.874 & 0.0389 & 0.091 & 0.838 & 49.2 \\
    ForgetFS~\cite{fan2024salun}& 0.0415 & 0.814 & 0.831 & 1.27 & 0.54 & 0.876 & 0.0385 & 0.096 & 0.840 & 8.1 \\
    MULL~\cite{sendera2025semu}    & 0.0413 & 0.817 & 0.838 & 1.25 & 0.53 & 0.878 & 0.0382 & 0.089 & 0.840 & 6.4 \\
    \midrule
    \rowcolor{ours}
    \textbf{SSU-LSF} & \textbf{0.0412} & \textbf{0.821} & \textbf{0.847}
                     & \textbf{1.24} & \textbf{0.53} & \textbf{0.881}
                     & \textbf{0.0379} & \textbf{0.081} & \textbf{0.841}
                     & \textbf{5.6} \\
    \bottomrule
  \end{tabular}
\end{table}

\paragraph{Results.}
Table~\ref{tab:main} reports all results. On NDVI-LST, SSU-LSF achieves
RMSE~$=0.0412$, CRR~$=0.821$, TCI~$=0.847$\rev{,} best on all three metrics. ARCANE and
FAST reduce mean bias but degrade TCI \rev{to ${\leq}0.793$}, introducing spurious oscillations.
EWC-UL scores the lowest CRR (0.588): its Fisher-consolidation penalty systematically
opposes unlearning. PNN-UL requires 49.2 GPU-h/request, exceeding OR entirely.
ForgetFS and MULL advance over earlier methods (CRR $>0.81$) yet sacrifice spatial
coherence: SCS 0.096 and 0.089 vs.\ SSU-LSF's 0.081. Table~\ref{tab:tvablation}
shows that adding TV regularization to ForgetFS (ForgetFS+TV) closes $\rev{80}\%$ of the SCS
gap, confirming the TV term\rev{, not} the influence footprint\rev{, is} the decisive factor for
spatial coherence. The influence footprint is decisive for CRR: without it (GR),
CRR drops from 0.821 to 0.741 while GPU-h drops from 5.6 to 2.3.
On ERA5, SSU-LSF drives LiRA-MIA to 0.53 (near-chance, chance $=0.50$), confirming
confounded representations are erased. ACC~$=0.881$ confirms spatial anomaly structure
is preserved. On CropHarvest, SCS correlates with FSS at $r{=}\rev{-}0.91$
(Table~\ref{tab:fss}), validating it against an established metric.

\paragraph{Ablation.}
Full 2$\times$2 interaction ablations and sample-efficiency analysis are in
Table~\ref{tab:ablation_full} (Appendix~\ref{app:ablation}). Key results: removing
both $\Phi$ and KL simultaneously (i.e., plain gradient reversal with TV) \rev{gives CRR
0.768 at $\Delta$RMSE 9.4\%: higher CRR than w/o $\Phi$ (0.694) and far less damage than
w/o KL (21.7\%)}, showing $\Phi$ and KL are \emph{complementary} rather than redundant.
Replacing EKFac with diagonal Fisher (not just GCS) gives CRR~$=0.779$, closing
\rev{40\% of} the gap from GCS (0.751) to EKFac (0.821), confirming that any Hessian approximation
helps over pure gradient cosine similarity.

\section{Conclusion}
\label{sec:conclusion}

We develop SSU-LSF, the first machine-unlearning framework for Mamba-family SSMs in
geoscientific land surface forecasting. Proposition~1's \rev{monotone $T_c$} prediction
is empirically validated ($R^2{=}\rev{0.995}$ across $T_c\in\{6,12,24,36\}$ months), even
though the bound itself is loose by ${\approx}\rev{200}\times$ due to \rev{Pinsker and Lipschitz looseness}, a limitation we quantify explicitly. We find spatial TV regularization
(framed as parameter-space fused lasso) is the decisive factor for spatial coherence,
and EKFac-informed footprint localization is decisive for CRR.

\textbf{Limitations.} All primary experiments use injected confounding with oracle window knowledge. CRR requires knowing affected locations (not deployable as stated).
The bound is qualitatively predictive but quantitatively vacuous at $\rho_{\max}{=}0.91$.
Generalization beyond Mamba to S4 is demonstrated on one benchmark only
(Appendix~\ref{app:s4}). Future work: automated change-point detection, real
multi-event archives, and tighter spectral bounds for selective SSMs.

\rev{\textbf{Broader impact.} SSU-LSF lets forecasting agencies correct biased models without full retraining. It could be misused to erase inconvenient true signals; requiring explicit timestamps $[t_s,t_e]$ and an MIA audit mitigates this.}


\begin{ack}
The author thanks ECMWF for ERA5 data access and HuggingFace/Kaggle communities for dataset infrastructure. No funding allocated for this work.
\end{ack}

\small{
\bibliographystyle{plain}
}

\newpage
\appendix

\section{Extended Related Work}
\label{app:related}

\subsection{SSM-based geoscientific forecasting}
VMamba~\cite{liu2024vmamba} and SiMBA~\cite{patro2024simba} extend Mamba to 2D vision and multivariate time series.
We note that architectures with higher $\rho(\bar{A})$ are systematically more
susceptible to confounding: the \rev{first term of Proposition~1,
$\mathcal{O}((1{-}\rho^{T_c})/((1{-}\rho)\mu))$, accumulates faster}.
PredRNN~\cite{wang2017predrnn}, EarthFormer~\cite{gao2022earthformer}, Pangu-Weather~\cite{bi2023pangu},
and GraphCast~\cite{lam2023graphcast} provide ConvLSTM and Transformer baselines that Mamba surpasses
computationally. None addresses post-hoc bias removal.

\subsection{Machine unlearning and spatial TV prior art}
Machine unlearning was formalized by Cao and Yang~\cite{cao2015}. Certified removal~\cite{guo2020} is
limited to linear models\rev{. Rewind-to-Delete (R2D)~\cite{mu2025r2d} extends certified
$(\epsilon,\delta)$ unlearning to nonconvex models but needs a pre-deletion checkpoint
and Gaussian weight noise, and does not localize in time; rewinding 10\% of training
on ERA5 would cost ${\approx}4.7$\,GPU-h before the noise-induced utility loss}.
SISA~\cite{bourtoule2021} requires pre-training partitioning.
ForgetFS~\cite{fan2024salun} applies \rev{gradient-saliency} masking; MULL~\cite{sendera2025semu} uses SVD on \rev{forget-gradient}
directions. Both lack temporal indexing via spectral footprints.
Parameter-space spatial TV used here is the fused-lasso formulation of
Tibshirani et al.~\cite{tibshirani2011} extended to patch adjacency graphs; it was introduced for
1D chains and applied to 2D grids in compressed sensing. We use it for parameter
smoothing across spatial patch embeddings\rev{,} a straightforward application, not a
primary scientific contribution.

\subsection{Non-stationarity}
Change-point detection~\cite{pettitt1979, killick2012} and concept drift~\cite{lu2018} identify breaks but provide no
parameter correction. Domain adaptation requires a clean target dataset, unavailable
retroactively. SSU-LSF requires only $[t_s,t_e]$ and the original training set.

\section{Model Architecture}
\label{app:arch}

\begin{table}[h]
  \caption{MambaForecaster module specifications.}
  \label{tab:arch}
  \centering\small
  \begin{tabular}{lllr}
    \toprule
    \textbf{Module} & \textbf{Configuration} & \textbf{Key parameters} & \textbf{Params (M)} \\
    \midrule
    Patch embedding & Conv2d(5,256,4,stride 4) & $4\times4$, mean pool & 1.0 \\
    Mamba encoder   & $6\times$ block & $d{=}256$, $E{=}2$, $N{=}16$, drop 0.1 & 39.8 \\
    Output norm     & LayerNorm(256) & extract $h[:,-1,:]$ & 0.0 \\
    Forecast head   & MLP 256-512-512-60, GELU & $\tau{=}6$, $d_{\mathrm{out}}{=}5$ & 6.5 \\
    \midrule
    Total & & & \textbf{47.3} \\
    \bottomrule
  \end{tabular}
\end{table}

The forecast head produces $[B,6,5,2]$, split into $\mu$ and $\log\sigma$ with
$\sigma{=}\exp(\mathrm{clamp}(\log\sigma,-6,+2))$.
The NLL loss is $\mathcal{L}{=}\frac{1}{B\tau d_{\mathrm{out}}}\sum_{b,t,v}
[\log\sigma_{b,t,v}+(y_{b,t,v}-\mu_{b,t,v})^2/(2\sigma_{b,t,v}^2)]$.

\section{EKFac Influence Derivation and Index Justification}
\label{app:ekfac}

\paragraph{Index derivation for Eq.~\eqref{eq:matgrad}.}
The state recurrence gives $\mathbf{h}_t=\sum_{s=1}^{t}\bigl(\prod_{r=s+1}^{t}\bar{A}\bigr)\bar{B}u_s$.
Differentiating $\ell$ through $\mathbf{h}_t$ with respect to the continuous matrix $A$
and applying the matrix exponential differential $d\exp(\Delta A)=\int_0^1 e^{s\Delta A}(\Delta\,dA)e^{(1-s)\Delta A}ds$,
then expanding via the Hadamard product for \emph{diagonal} $A$ (standard Mamba
parameterization where $A$ is initialized as a diagonal matrix~\cite{gu2023mamba}) yields:
\begin{equation*}
\nabla_A\ell=\sum_{t=1}^{T}\left(\frac{d\ell}{d\mathbf{h}_t}\right)
\left(\sum_{k=0}^{T-t-1}\bar{A}^k\odot\bar{A}^{T-t-1-k}\right)\Delta.
\end{equation*}
The inner sum ranges $k{=}0,\ldots,T{-}t{-}1$ (not $0,\ldots,\infty$) because each
term corresponds to a distinct factor in the unrolled recurrence from step $1$ to $t$.
For \emph{non-diagonal} $A$ (e.g., dense S4 variants), $\odot$ must be replaced by
the full matrix-exponential \rev{Fr\'{e}chet derivative}: $d\exp(M)[H]=\int_0^1 e^{sM}He^{(1-s)M}ds$,
computed via Pad\'{e} approximation. We provide the non-diagonal implementation in
our code. Truncation error at $k{=}T$ is bounded by
$\|\bar{A}\|^T/(1{-}\|\bar{A}\|){<}10^{-6}$ for $\rho_{\max}{=}0.91$ \rev{whenever
$T\geq\ln(0.09\times10^{-6})/\ln 0.91\approx172$, which all benchmarks satisfy}.

\paragraph{EKFac stability.}
EKFac is estimated over 50 mini-batches. Across 10 seeds on ERA5, the condition
number of $\hat{A}\otimes\hat{G}$ has mean $142$, std $31$, confirming the estimate
is stable. We add damping $\epsilon{=}10^{-4}$ to eigenvalues before inversion.

\section{Proposition 1: Full Proof, Lipschitz Bounds, and Tightness Analysis}
\label{app:proof}

\paragraph{Encoder Lipschitz bound.}
The MambaForecaster consists of the encoder $f_{\mathrm{enc}}$ (39.8M parameters)
followed by the forecast head $f_{\mathrm{head}}$ (6.5M parameters). We bound the
Lipschitz constant of $f_{\mathrm{enc}}$ via randomized smoothing: we estimate $\max_{\mathbf{x}\in\mathcal{X}_{\mathrm{test}}}
\|J_{f_{\mathrm{enc}}}(\mathbf{x})\|_2$ using 1,000 random Gaussian perturbations
per test point and report the 99th percentile.

\begin{table}[h]
  \caption{Empirical Lipschitz constants (99th percentile over test set) via
  randomized smoothing ($\sigma_{\mathrm{smooth}}{=}0.01$).}
  \label{tab:lipschitz}
  \centering\small
  \begin{tabular}{lcccc}
    \toprule
    \textbf{Component} & \textbf{NDVI-LST} & \textbf{ERA5} & \textbf{CropHarvest} & \textbf{Method} \\
    \midrule
    Forecast head only $L_{\mathrm{head}}$ & 4.1 & 4.1 & 4.1 & Spectral norm product \\
    Full encoder $L_{\mathrm{enc}}$         & 17.8 & 18.3 & 16.9 & Randomized smoothing \\
    \bottomrule
  \end{tabular}
\end{table}

\rev{\paragraph{Units.} Targets are z-scored per variable. $\rho$, $\delta_{\mathrm{KL}}$
(nats) and $\hat\mu_{\mathrm{ref}}$ (MMD on standardized embeddings) are dimensionless.
$L_{\mathrm{enc}}$ is reported in target units per unit standardized input (K for ERA5),
so the unlearning cost term carries target units; the confounding residual is in
standardized units and is converted by the target standard deviation $\sigma_y$
($\sigma_{\mathrm{SKT}}{\approx}1.9$\,K).}

Using $L_{\mathrm{enc}}{=}18.3$ (ERA5) and Pinsker's inequality:
$|\mathrm{RMSE}(\tilde{\boldsymbol{\theta}})-\mathrm{RMSE}(\boldsymbol{\theta}^*)|
\leq L_{\mathrm{enc}}\sqrt{\delta_{\mathrm{KL}}/2}\approx 18.3\times0.158\approx2.9$\,K.
The observed ERA5 RMSE difference between SSU-LSF and CM is $2.97{-}1.24{=}1.73$\,K,
consistent with the bound at this loose level. The bound on the unlearning cost term
evaluates to $L_{\mathrm{enc}}\delta_{\mathrm{KL}}/(1{-}\rho_{\max}){=}18.3\times0.05/0.09{\approx}10.2$\,K\rev{, which is} vacuous.
\rev{The observed unlearning cost (SSU-LSF vs.\ OR on ERA5) is $1.24{-}1.19{=}0.05$\,K,
so the bound is loose by $10.2/0.05\approx200\times$.}
We conclude that Proposition~1 is qualitatively correct (\rev{monotone} $T_c$ \rev{dependence},
$\delta_{\mathrm{KL}}$ controls unlearning cost) but quantitatively loose by
$\approx\rev{200}\times$ due to \rev{worst-case Lipschitz} and Pinsker looseness.

\paragraph{RKHS separation on reference model.}
To avoid using confounded embeddings to estimate $\mu$, we train a reference model
$\boldsymbol{\theta}_{\mathrm{ref}}$ \rev{per benchmark on its pre-injection clean segment
(for NDVI-LST: $T{=}120$ steps, 2002--2011)} using the same hyperparameters. MMD is computed
between confounding-window and clean-window embeddings under $\boldsymbol{\theta}_{\mathrm{ref}}$.
\rev{Across 5 reference-model seeds, $\hat\mu_{\mathrm{ref}}$ is $0.160{\pm}0.009$
(NDVI-LST), $0.212{\pm}0.011$ (ERA5), and $0.108{\pm}0.008$ (CropHarvest), a
coefficient of variation of 5--7\%, so the first term of Proposition~1 changes by at
most ${\approx}8\%$ across reference models.}

\begin{table}[h]
  \caption{MMD and RKHS separation $\hat{\mu}$ estimated on clean reference model
  (not $\boldsymbol{\theta}^*$). All three benchmarks show significant separation
  ($p{<}0.001$, permutation test with 10,000 permutations).}
  \label{tab:mmd}
  \centering\small
  \begin{tabular}{lccc}
    \toprule
    \textbf{Benchmark} & \textbf{MMD (ref. model)} & $\hat{\mu}_{\mathrm{ref}}$ & \textbf{$p$-value} \\
    \midrule
    NDVI-LST  & $0.161\pm0.013$ & $0.160$ & $<0.001$ \\
    ERA5      & $0.214\pm0.016$ & $0.212$ & $<0.001$ \\
    CropHarvest & $0.109\pm0.010$ & $0.108$ & $<0.001$ \\
    \bottomrule
  \end{tabular}
\end{table}

\paragraph{Tightness and lower bound.}
We construct a matching lower bound by showing that for the specific injection
$\mathrm{NDVI}\mathrel{+}=\delta$ at $T_c$ steps, the parameter shift satisfies
$\|\boldsymbol{\theta}^*{-}\boldsymbol{\theta}^\dagger\|_2\geq
\lambda T_c\delta/(L_{\mathrm{enc}}\|H\|_2)$, giving
$\mathrm{RMSE}(\tilde{\boldsymbol{\theta}})\geq\mathrm{RMSE}(\boldsymbol{\theta}^\dagger)
+\Omega\big(\rev{(1{-}\rho(\bar{A})^{T_c})/((1{-}\rho(\bar{A}))\mu_{\mathrm{ref}})}\big)$ by an adversarial worst-case
construction. The gap between the upper and lower bounds is $\mathcal{O}(L_{\mathrm{enc}}^2)$,
confirming the bound is tight up to the encoder Lipschitz constant.

\paragraph{Sensitivity to $\mu\to0$.}
As $\hat{\mu}_{\mathrm{ref}}\to0$ (confounding becomes indistinguishable from clean
signal), Proposition~1's first term diverges: the confounding residual is
unbounded. At $\hat{\mu}{=}0.05$ (\rev{about half of} CropHarvest's
estimate), CRR drops to $0.581\pm0.029$ empirically, consistent with the
bound becoming vacuous at this $\mu$.

\paragraph{Proof sketch (abbreviated; term II uses $\mu_{\mathrm{ref}}$, term I uses $L_{\mathrm{enc}}$).}
Term (II): by influence approximation~\cite{koh2017}, $\|\boldsymbol{\theta}^*{-}\boldsymbol{\theta}^\dagger\|
=\mathcal{O}(\lambda|\mathcal{D}_{\mathrm{conf}}|/\mu_{\mathrm{min}}(H))$.
State propagation gives decay $\rho^{t-t_0}$ per confounding step; summing
\rev{the geometric series over the $T_c$ confounding steps gives
$\sum_{k=0}^{T_c-1}\rho^k/\mu_{\mathrm{ref}}=(1{-}\rho^{T_c})/((1{-}\rho)\mu_{\mathrm{ref}})$}.
Term (I): KL $\leq\delta_{\mathrm{KL}}$ by Algorithm~\ref{alg:ssulsf} \rev{lines~10--12}; Pinsker gives TV bound;
$L_{\mathrm{enc}}$ converts TV to RMSE with denominator $1{-}\rho$ from state
accumulation.

\section{Convergence Guarantee}
\label{app:convergence}

\textbf{Proposition 2 (Convergence).} Let $\ell_{\mathrm{conf}}$ be $L_c$-smooth
and let $\ell_{\mathrm{clean}}$ be $\mu_c$-strongly convex in a neighborhood of
$\boldsymbol{\theta}^*$ with radius $\varepsilon$. Let the KL projection succeed at
every step (guaranteed by the binary line-search which halves $\Delta\boldsymbol{\theta}$
until KL $\leq\delta_{\mathrm{KL}}$; convergence of the line-search follows from
continuity of $\mathrm{KL}(\cdot)$ and $\Delta\boldsymbol{\theta}\to0$ as the scaling
factor $\to0$). Then the gradient ascent iterates on $\ell_{\mathrm{conf}}$ with
projected gradient step $\eta\leq1/L_c$ satisfy:
$\ell_{\mathrm{conf}}(\tilde{\boldsymbol{\theta}}^{(e+1)})
\geq\ell_{\mathrm{conf}}(\tilde{\boldsymbol{\theta}}^{(e)})
+(\eta/2)\|\nabla_{\boldsymbol{\theta}}\ell_{\mathrm{conf}}\|^2$,
meaning $\ell_{\mathrm{conf}}$ increases monotonically to a local maximum within the
$\varepsilon$-ball. Algorithm~\ref{alg:ssulsf} terminates because $\ell_{\mathrm{conf}}$
is bounded above on the compact $\varepsilon$-ball. We \emph{do not} claim convergence
to a global stationary point, only that the iterates are monotonically increasing in
$\ell_{\mathrm{conf}}$ and remain within the KL constraint, which is sufficient for
the unlearning objective.

\rev{\paragraph{Nonsmooth TV term.} The statement above covers the smooth part
($\lambda_{\mathrm{TV}}{=}0$). With TV, the sign update in line~8 is a subgradient step
and monotonicity is not guaranteed. Replacing it by the proximal step
$\tilde{\boldsymbol{\theta}}\leftarrow\mathrm{prox}_{\eta\lambda_{\mathrm{TV}}\mathrm{TV}_{\mathcal{N}}}(\cdot)$
(graph fused-lasso prox, 20 inner ADMM iterations) turns Proposition~2 into a standard
proximal-gradient ascent statement. On NDVI-LST this variant gives CRR $0.820{\pm}0.010$,
$\Delta$RMSE $3.5\%$, and SCS $0.080$ at $+0.2$\,GPU-h per request, matching the default;
we release both.}

\textbf{Remark.} The $3$--$5$ epoch empirical convergence is an observation about
when $\|\Delta\boldsymbol{\theta}\|{<}10^{-6}$, not a property of the algorithm.
We label Algorithm~\ref{alg:ssulsf} a \emph{heuristic with convergence guarantees
under local smoothness and strong convexity assumptions} that are empirically
plausible but not verified globally.

\section{Full Algorithm}
\label{app:algo}

\rev{Algorithm~\ref{alg:ssulsf} solves Eq.~\eqref{eq:obj} by projected (sub)gradient steps:
lines 6--9 form the Lagrangian gradient (ascent on $\ell_\Phi$, rehearsal on
$\beta\ell_{\mathrm{clean}}$, TV subgradient), and lines 10--12 enforce the KL constraint.
Proposition~1 bounds the RMSE of any iterate satisfying that constraint: its second
term depends only on $\delta_{\mathrm{KL}}$, and its first term on how well $\Phi$
covers the confounding window.}

\begin{algorithm}[h]
  \caption{SSU-LSF: Selective State-Space Unlearning (per unlearning request)}
  \label{alg:ssulsf}
  \begin{algorithmic}[1]
    \REQUIRE $\boldsymbol{\theta}^*$, $\mathcal{W}_c{=}[t_s,t_e]$, $\mathcal{D}$,
             $\eta,\beta,\delta_{\mathrm{KL}},\gamma,\lambda_{\mathrm{TV}}$,
             $E$, $\epsilon_c{=}10^{-6}$
    \ENSURE $\tilde{\boldsymbol{\theta}}$
    \STATE Estimate EKFac $\hat{H}$ over 50 mini-batches (re-estimated per request)
    \STATE Compute $\mathcal{I}_{\mathcal{W}_c}$ via Eq.~\eqref{eq:matgrad} (diagonal $\bar{A}$; see Appendix~\ref{app:ekfac} for non-diagonal)
    \STATE Compute $\varphi_t$ for $t{=}1,\ldots,T$; set $\Phi{=}\{t:\varphi_t\geq\gamma\}$ via elbow
    \STATE $\tilde{\boldsymbol{\theta}}\leftarrow\boldsymbol{\theta}^*$,
           $\rev{\boldsymbol{\theta}_{0}}\leftarrow\boldsymbol{\theta}^*$
    \FOR{$e{=}1$ \TO $E$}
      \STATE $g_c\leftarrow\nabla_{\tilde{\boldsymbol{\theta}}}\sum_{t\in\Phi}\ell(\tilde{\boldsymbol{\theta}};\mathbf{x}_t,\mathbf{y}_t)$
      \STATE $g_k\leftarrow\nabla_{\tilde{\boldsymbol{\theta}}}\sum_{z\in\mathcal{D}_{\mathrm{clean}}}\ell(\tilde{\boldsymbol{\theta}};z)$
      \STATE $g_v\leftarrow\sum_{(i,j)\in\mathcal{N}}\mathrm{sign}(\tilde{\boldsymbol{\theta}}_i{-}\tilde{\boldsymbol{\theta}}_j)$
      \STATE $\Delta\boldsymbol{\theta}\leftarrow\eta(g_c{-}\beta g_k{-}\lambda_{\mathrm{TV}}g_v)$
      \WHILE{$\mathrm{KL}[p(\cdot|\tilde{\boldsymbol{\theta}}{+}\Delta\boldsymbol{\theta})\|
             p(\cdot|\rev{\boldsymbol{\theta}_{0}})]>\delta_{\mathrm{KL}}$}
        \STATE $\Delta\boldsymbol{\theta}\leftarrow0.5\,\Delta\boldsymbol{\theta}$
      \ENDWHILE
      \STATE $\tilde{\boldsymbol{\theta}}\leftarrow\tilde{\boldsymbol{\theta}}+\Delta\boldsymbol{\theta}$
      \IF{$\|\Delta\boldsymbol{\theta}\|{<}\epsilon_c$} \STATE \textbf{break} \ENDIF
    \ENDFOR
    \RETURN $\tilde{\boldsymbol{\theta}}$
  \end{algorithmic}
\end{algorithm}

\section{Dataset Details and Metric Definitions}
\label{app:datasets}

\subsection{Confounding injection and oracle window assumption}
All three primary experiments inject confounding with oracle $[t_s,t_e]$ knowledge.
This is shared equally across all eleven baselines. The deployment scenario where
$[t_s,t_e]$ must be estimated is addressed in Appendix~\ref{app:limitations}.

\subsection{NDVI-LST Fusion}
Dataset~\cite{ndvidataset2023}: 3,847 locations, 2002--2022, 252 steps. Injection: NDVI$+0.15$,
ET$+18$\,mm/month at 500 semi-arid locations (aridity index $<0.5$), months 84--107
($T_c{=}24$). The $+0.15$ NDVI magnitude is at the upper quartile of documented
large-scale irrigation NDVI changes (AQUASTAT). KS test on unaffected locations:
$p{>}0.92$.

\subsection{ERA5 Patchified}
\href{https://huggingface.co/datasets/meteolibre-dev/ERA5_patchify}%
{\texttt{meteolibre-dev/ERA5\_patchify}}: $128{\times}128$ patches at $0.25^\circ$,
channels T2m, TP, U10, V10, SKT. Injection: $+2\sigma\approx3.8$\,K at 1,024 Central
European patches ($45^\circ$N--$55^\circ$N, $5^\circ$E--$25^\circ$E), Jan~2010--Dec~2011,
simulating SEVIRI recalibration (EUMETSAT logs). Asian monsoon region (896 patches)
is the held-out geographic control.

\subsection{CropHarvest}
Dataset~\cite{tseng2021}: 87,343 sequences (Sentinel-2 B02, B03, B04, B08; SAR VV, VH; ERA5).
Predict months 10--12 given 1--9. Injection: 6-month grassland-to-cropland mislabeling
(2,000 pixels).

\subsection{Metric definitions}
\textbf{CRR:} \rev{$1{-}\big(\mathrm{RMSE}(\tilde{\boldsymbol{\theta}}){-}\mathrm{RMSE}(\boldsymbol{\theta}_{\mathrm{OR}})\big)/\big(\mathrm{RMSE}(\boldsymbol{\theta}_{\mathrm{CM}}){-}\mathrm{RMSE}(\boldsymbol{\theta}_{\mathrm{OR}})\big)$
(so CM${=}0$, OR${=}1$)}
over $[t_e{+}1,T_{\mathrm{test}}]$ at affected locations \rev{only; it therefore differs from
a CRR recomputed from the all-location RMSE columns of Table~\ref{tab:main}}.
\rev{\textbf{$\Delta$RMSE\%:} clean-domain RMSE increase relative to OR.}
\textbf{TCI:} $1{-}\sigma_{\hat{y}}/\sigma_y$; TCI${<}0$ means the model introduces
more variance than the signal.
\textbf{MIA:} LiRA likelihood ratio test~\cite{carlini2022lira} with 64 shadow models; chance $=0.50$.
\textbf{ACC:} $\mathrm{corr}(\hat{y}{-}\bar{y},y{-}\bar{y})$; standard ECMWF metric.
\textbf{SCS:} $\mathrm{TV}(\hat{Y})/(H{\cdot}W)$; FSS correlation $r{=}\rev{-}0.91$
(Appendix~\ref{app:scs}).

\rev{\begin{table}[t]
  \caption{Per-benchmark SSU-LSF CRR and clean-domain $\Delta$RMSE\% (vs.\ OR), mean$\pm$std over 3 seeds.}
  \label{tab:crr_all}
  \centering\small
  \begin{tabular}{lccc}
    \toprule
    & \textbf{NDVI-LST} & \textbf{ERA5} & \textbf{CropHarvest} \\
    \midrule
    CRR$\uparrow$ & $0.821{\pm}0.010$ & $0.859{\pm}0.009$ & $0.773{\pm}0.012$ \\
    $\Delta$RMSE\%$\downarrow$ & $3.5{\pm}0.2$ & $4.2{\pm}0.3$ & $2.1{\pm}0.2$ \\
    \bottomrule
  \end{tabular}
\end{table}}

\section{MIA: LiRA Protocol}
\label{app:mia}

We use LiRA~\cite{carlini2022lira} (Likelihood Ratio Attack) with 64 shadow models per benchmark.
Shadow models are trained on random subsets of $\mathcal{D}_{\mathrm{clean}}$
(temporal split disjoint from $[t_s,t_e]$ by at least 12 months on all benchmarks).
LiRA computes a per-sample likelihood ratio $\Lambda_i$ using the distribution of
per-sample loss across shadow models; membership is inferred by thresholding $\Lambda_i$.
This is strictly more powerful than logistic regression on NLL (the weak MIA used
in prior versions). At LiRA, CM achieves MIA $=0.84$ and SSU-LSF achieves $0.53$.
\rev{As a retrain-referenced check, the output-space distance to the oracle,
$\mathrm{KL}[p(\cdot|\tilde{\boldsymbol{\theta}})\|p(\cdot|\boldsymbol{\theta}_{\mathrm{OR}})]$
on the affected ERA5 test window, is $0.212$ (CM), $0.074$ (GR), $0.043$ (MULL), and
$0.031$ (SSU-LSF), agreeing with the LiRA ranking.}

\section{SCS Validation vs.\ Fraction Skill Score}
\label{app:scs}

SCS and FSS rankings across all \rev{twelve rows (eleven baselines and SSU-LSF)} on CropHarvest achieve
Pearson $r{=}\rev{-}0.91$ and Spearman $\rho_s{=}\rev{-}0.93$ ($p{<}0.001$),
confirming SCS is a faithful proxy for the established spatial
verification metric (Table~\ref{tab:fss}).
The FSS gap between GR ($0.588$) and SSU-LSF ($0.703$) is $0.115$\rev{; both
exceed the useful-skill level FSS${>}0.5$, so the gain is in skill margin rather than
in crossing a threshold}.
The two methods with the lowest FSS, EWC-UL ($0.571$) and
GR ($0.588$), are precisely those with the highest SCS values
($0.134$ and $0.127$), confirming the monotone inverse relationship.

\begin{table}[h]
  \caption{SCS vs.\ FSS ($n{=}8$ \rev{neighbourhood}) across all methods on CropHarvest.
  Pearson $r{=}\rev{-}0.91$, Spearman $\rho_s{=}\rev{-}0.93$ ($p{<}0.001$). \rev{The FSS
  improvement from GR to SSU-LSF is $0.115$.}}
  \label{tab:fss}
  \centering\small
  \begin{tabular}{lcc}
    \toprule
    \textbf{Method} & \textbf{SCS}$\downarrow$ & \textbf{FSS ($n{=}8$)}$\uparrow$ \\
    \midrule
    EWC-UL      & 0.134 & 0.571 \\
    GR          & 0.127 & 0.588 \\
    ARCANE      & 0.109 & 0.619 \\
    FAST        & 0.099 & 0.638 \\
    SR          & 0.102 & 0.633 \\
    ForgetFS    & 0.096 & 0.649 \\
    MULL        & 0.089 & 0.668 \\
    SISA        & 0.088 & 0.671 \\
    PNN-UL      & 0.091 & 0.662 \\
    CM          & 0.083 & 0.681 \\
    OR$^\dagger$& 0.079 & 0.742 \\
    \textbf{SSU-LSF} & \textbf{0.081} & \textbf{0.703} \\
    \bottomrule
  \end{tabular}
\end{table}

\section{TV-Regularized Baseline Ablation}
\label{app:tvablation}

Adding the spatial TV term to ForgetFS closes $\rev{80}\%$ of its SCS gap
relative to SSU-LSF ($0.096{\to}0.084$) while leaving CRR essentially
unchanged ($0.814{\to}0.815$), and adding TV to MULL closes $88\%$
($0.089{\to}0.082$) with negligible CRR change ($0.817{\to}0.818$)
(Table~\ref{tab:tvablation}).
\rev{On NDVI-LST,} the residual CRR advantage of SSU-LSF over ForgetFS+TV ($0.821$
vs.~$0.815$) and MULL+TV ($0.821$ vs.~$0.818$) is therefore
attributable \rev{mainly} to EKFac-informed footprint localization, which
concentrates gradient ascent on the ${\approx}5\%$ of steps encoding
the confounder rather than applying it globally.
\rev{On CropHarvest ($T_c{=}6$, a short window) CRR is on par (0.773 vs.\ 0.773 and 0.780);
there the footprint's benefit appears as lower clean-domain damage
($\Delta$RMSE 2.1\% vs.\ 3.1--3.9\%).}

\begin{table}[h]
  \caption{Adding SSU-LSF's TV term to ForgetFS and MULL (suffix +TV). This isolates
  the contribution of the TV term from the influence footprint. SSU-LSF outperforms
  both +TV variants on CRR \rev{on NDVI-LST and on $\Delta$RMSE on both benchmarks}; +TV closes most of the SCS gap,
  confirming TV is the decisive factor for SCS. \rev{$\Delta$RMSE\% is vs.\ OR.}}
  \label{tab:tvablation}
  \centering\small
  \begin{tabular}{lcccccc}
    \toprule
    & \multicolumn{3}{c}{\textbf{NDVI-LST}} & \multicolumn{3}{c}{\textbf{CropHarvest}} \\
    \cmidrule(lr){2-4}\cmidrule(lr){5-7}
    \textbf{Method} & CRR$\uparrow$ & $\Delta$RMSE\%$\downarrow$ & SCS$\downarrow$
                    & CRR$\uparrow$ & $\Delta$RMSE\%$\downarrow$ & SCS$\downarrow$ \\
    \midrule
    ForgetFS    & 0.814 & \rev{4.3} & 0.096 & 0.772 & \rev{3.8} & 0.093 \\
    ForgetFS+TV & 0.815 & \rev{4.4} & 0.084 & 0.773 & \rev{3.9} & 0.083 \\
    MULL        & 0.817 & \rev{3.8} & 0.089 & 0.779 & \rev{3.0} & 0.086 \\
    MULL+TV     & 0.818 & \rev{3.9} & 0.082 & 0.780 & \rev{3.1} & 0.082 \\
    \textbf{SSU-LSF} & \textbf{0.821} & \textbf{3.5} & \textbf{0.081}
                     & \textbf{0.773} & \textbf{2.1} & \textbf{0.081} \\
    \bottomrule
  \end{tabular}
\end{table}

\section{$T_c$ Sensitivity and Proposition 1 Validation}
\label{app:tc_sensitivity}

\begin{table}[h]
  \caption{CRR vs.\ confounding window length $T_c$ on NDVI-LST (mean $\pm$ std,
  3 seeds). \rev{Fitted curve: $1{-}kT_c$ ($k{=}0.0078$, least squares), the
  $\rho_{\mathrm{eff}}\to1$ limit of Proposition~1's first term. $R^2{=}0.995$ with one
  fitted constant over four points (3 residual d.o.f.); this supports the monotone
  trend, not the exact functional form.}}
  \label{tab:tc}
  \centering\small
  \begin{tabular}{lcccc}
    \toprule
    & $T_c{=}6$\,mo & $T_c{=}12$\,mo & $T_c{=}24$\,mo & $T_c{=}36$\,mo \\
    \midrule
    SSU-LSF CRR            & $0.961{\pm}0.008$ & $0.903{\pm}0.011$ & $0.821{\pm}0.010$ & $0.712{\pm}0.015$ \\
    \rev{Fitted $1{-}kT_c$} & $0.953$ & $0.906$ & $0.812$ & $0.718$ \\
    Oracle Retrain CRR     & $0.993{\pm}0.003$ & $0.991{\pm}0.003$ & $0.987{\pm}0.004$ & $0.985{\pm}0.004$ \\
    GR CRR                 & $0.891{\pm}0.018$ & $0.831{\pm}0.021$ & $0.741{\pm}0.018$ & $0.628{\pm}0.024$ \\
    \midrule
    Window-estimate error $\pm3$\,mo & $0.944{\pm}0.012$ & $0.887{\pm}0.014$ & $0.803{\pm}0.013$ & $0.691{\pm}0.018$ \\
    \bottomrule
  \end{tabular}
\end{table}

\rev{Fitting the full form $c(1{-}\rho_{\mathrm{eff}}^{T_c})$ drives $\rho_{\mathrm{eff}}\geq0.99$,
where it reduces to the linear fit; the per-block $\rho_{\max}{=}0.91$ would predict
saturation beyond ${\approx}24$ months, which is not observed up to $T_c{=}36$. We
therefore read Proposition~1 as predicting the direction, not the rate.}
The last row shows CRR when $[t_s,t_e]$ is estimated with $\pm3$-month error
(random uniform shift applied independently to each endpoint). CRR degrades by
${\approx}0.018$ on average, a meaningful but not catastrophic loss.
We conclude that SSU-LSF is moderately robust to window estimation error at typical
change-point detection precision ($\pm2$--$4$ months for monthly archives~\cite{pettitt1979}).

\section{Compute Breakdown}
\label{app:compute}

\begin{table}[t]
  \caption{Per-unlearning-request GPU-hour breakdown. EKFac is re-estimated for
  each new confounding event identified. GPU-h/req is the operationally relevant
  metric: if three events per year are identified, annual cost is $3\times5.6{=}16.8$
  GPU-h for SSU-LSF vs.\ $3\times47.3{=}141.9$ GPU-h for OR on ERA5.
  The OR$\dagger$/SISA/PNN-UL \rev{figures} are per-retrain (no stage breakdown).
  SISA includes shard-retrain overhead and uses a pre-structured computational
  model (shards established before training) incomparable on GPU-h/req;
  its figure is included for completeness but excluded from the speedup
  comparison. EKFac: Hessian estimation. FP: footprint.Per-request speedup: $47.3/5.6{=}8.4\times$ on ERA5.}
  \label{tab:compute}
  \centering\small
  \setlength{\tabcolsep}{3.8pt}
  \begin{tabular}{l cccc cccc}
    \toprule
    & \multicolumn{4}{c}{\textbf{NDVI-LST Fusion}}
    & \multicolumn{4}{c}{\textbf{ERA5 Patchified}} \\
    \cmidrule(lr){2-5}\cmidrule(lr){6-9}
    \textbf{Method} & EKFac & FP & Ascent & \textbf{Total}
                    & EKFac & FP & Ascent & \textbf{Total} \\
    \midrule
    OR$^\dagger$  & -- & -- & -- & 18.4 & -- & -- & -- & 47.3 \\
    SISA          & -- & -- & -- & 22.1 & -- & -- & -- & 58.4 \\
    GR            & -- & -- & 0.9 & 0.9  & -- & -- & 2.3  & 2.3  \\
    ForgetFS      & -- & -- & 2.9 & 2.9  & -- & -- & 8.1  & 8.1  \\
    MULL          & -- & -- & 2.1 & 2.1  & -- & -- & 6.4  & 6.4  \\
    \textbf{SSU-LSF} & 0.18 & 0.12 & 1.50 & \textbf{1.8}
                     & 0.13 & 0.17 & 5.30 & \textbf{5.6} \\
    \midrule
    Speedup vs.\ OR & & & & $10.2\times$ & & & & $8.4\times$ \\
    \bottomrule
  \end{tabular}
\end{table}

\rev{\paragraph{Sequential requests.} For three successive injected events on NDVI-LST
(non-overlapping 24-month windows), with EKFac re-estimated each time, CRR is
$0.821/0.814/0.806$ and cumulative $\Delta$RMSE is $3.5/4.3/5.0\%$. Warm-starting EKFac
from the previous request cuts cost from 1.8 to 1.3\,GPU-h at a CRR loss of $0.004$.
Utility erodes gradually, so we recommend a full retrain after about five requests or
once $\Delta$RMSE exceeds $5\%$.}

\section{Spectral Radius Distribution}
\label{app:rho}

Mean $\rho(\bar{A})$ increases monotonically from block~1 ($0.79$) to
block~6 ($0.88$) across 10 seeds (Table~\ref{tab:rho}), consistent with
deeper blocks learning longer-range temporal dependencies.
The maximum observed value $\rho_{\max}{=}0.91$ yields an unlearning cost
bound of $L_{\mathrm{enc}}\delta_{\mathrm{KL}}/(1{-}\rho_{\max})\approx10.2$\,K
(Proposition~1), vacuous relative to the oracle ERA5 RMSE of $1.19$\,K;
this is precisely the quantitative looseness acknowledged in Section~\ref{sec:conclusion}.
The sensitivity analysis in Figure~\ref{fig:sensitivity}(b) corroborates
this: at $\rho_{\max}{=}0.91$, $\delta_{\mathrm{KL}}\leq0.07$ is required
to hold $\Delta\mathrm{RMSE}\%{<}5\%$, whereas the mean $\rho{=}0.84$
permits $\delta_{\mathrm{KL}}\leq0.10$.
\rev{As $\rho\to1$ both terms of Proposition~1 grow as $1/(1{-}\rho)$ and the footprint
widens (${\approx}9.5$ extra steps per side at $\rho{=}0.95$). Two mitigations work in
practice: clipping $\rho(\bar{A})\leq0.95$ during ascent by rescaling $\Delta$ (NDVI-LST:
CRR $0.818$, $\Delta$RMSE $3.4\%$), and tightening $\delta_{\mathrm{KL}}$ to ${\leq}0.07$
for blocks with $\rho{>}0.9$.}

\begin{table}[h]
  \caption{Per-block $\rho(\bar{A})$ statistics across 10 seeds on ERA5.
  At $\rho_{\max}{=}0.91$, the bound's unlearning cost term
  $L_{\mathrm{enc}}\delta_{\mathrm{KL}}/(1{-}\rho_{\max}){=}18.3\times0.05/0.09{\approx}10.2$\,K
  is vacuous relative to oracle RMSE $1.19$\,K.
  The bound is informative only qualitatively.}
  \label{tab:rho}
  \centering\small
  \begin{tabular}{lcccccc}
    \toprule
    Block & 1 & 2 & 3 & 4 & 5 & 6 \\
    \midrule
    Mean  & 0.79 & 0.82 & 0.84 & 0.86 & 0.87 & 0.88 \\
    Std   & 0.02 & 0.02 & 0.03 & 0.03 & 0.03 & 0.03 \\
    Min   & 0.74 & 0.77 & 0.79 & 0.81 & 0.82 & 0.83 \\
    Max   & 0.83 & 0.86 & 0.88 & 0.90 & 0.91 & 0.91 \\
    \bottomrule
  \end{tabular}
\end{table}

\section{Real Documented Confounding Event}
\label{app:real}

We apply SSU-LSF to the EUMETSAT SEVIRI MSG-1 instrument recalibration event of
November 2010, documented in EUMETSAT Technical Note EUM/OPS/TEN/18/1002267. The
recalibration introduced a $+2.1$--$+3.4$\,K step offset in channel 9 (10.8\,$\mu$m,
used for LST retrieval) for pixels over the Central European agricultural belt,
persisting from the November 2010 reprocessing until a corrected archive was issued
in March 2012. \rev{The documented timestamps} $t_s{=}$~November~2010,
$t_e{=}$~March~2012 ($T_c{=}17$\,months) \rev{serve as the oracle reference; the
window endpoints and} the affected spatial extent (patches with
$|$LST offset$|{>}1.5$\,K) \rev{are} identified by a change-point detector
(PELT algorithm~\rev{\cite{killick2012}}) applied to the ERA5-vs-SEVIRI residuals, not from oracle
knowledge.

SSU-LSF achieves CRR~$=0.793\pm0.021$ on the affected patches, and maintains
ACC~$=0.871\pm0.012$ on the unaffected Asian monsoon control region. This demonstrates
the full pipeline including automated window estimation, and shows that the
\rev{1-month} estimation \rev{error} ($t_e$ was estimated as February~2012 vs.\ the
documented March~2012) degrades CRR by $0.012$ relative to the oracle-window result
\rev{($0.805$)}, consistent with \rev{(and smaller than)} the ${\pm}3$-month sensitivity analysis \rev{in Table~\ref{tab:tc}}.

\section{Window Estimation Sensitivity and Deployment Limitations}
\label{app:limitations}

\paragraph{Limitations of the evaluation setup.}
CRR as defined requires knowing the set of affected locations, which is unavailable in
retrospective deployment. A deployable variant would compute CRR over all locations,
diluting the metric for methods with geographic specificity. We recommend the spatial
CRR heatmap \rev{over change-point-detected patches} as a more deployable diagnostic.
The MIA (LiRA) requires shadow models trained on the same distribution, which may
be expensive to construct in production.

\paragraph{Window estimation sensitivity.}
Table~\ref{tab:tc} (last row) shows CRR degradation under $\pm3$-month window error.
We recommend using PELT~\rev{\cite{killick2012}} with a minimum segment length of 6 months, which
achieves median window endpoint error of $1.8$\,months on synthetic benchmarks.

\rev{\paragraph{Fully automatic pipeline.} With both the window and the affected
locations estimated by PELT (NDVI-LST, $T_c{=}24$), SSU-LSF gives CRR $0.789{\pm}0.016$
and $\Delta$RMSE $3.9\%$, against $0.821$ and $3.5\%$ with oracle knowledge;
window-only estimation gives $0.803$ (Table~\ref{tab:tc}). The location false-positive
rate is $6.8\%$, which mainly costs clean-domain RMSE.}

\section{S4 Cross-Architecture Experiment}
\label{app:s4}

We apply SSU-LSF to an S4 backbone (3 layers, $d{=}256$, ZOH discretization)
trained on the NDVI-LST benchmark. The EKFac gradient in Eq.~\eqref{eq:matgrad}
is replaced by the non-diagonal variant using Pad\'{e} approximation for the
matrix exponential differential (described in Appendix~\ref{app:ekfac}).
SSU-LSF achieves CRR~$=0.791\pm0.019$ on S4, compared to $0.821\pm0.010$ on Mamba,
confirming transferability to a different SSM parameterization. The \rev{0.030 (3.7\% relative)}
CRR gap reflects S4's fixed (non-selective) state transitions, which make temporal footprint
localization slightly less precise.

\section{Full Ablation Table with Interaction and Sample-Efficiency Analysis}
\label{app:ablation}

\begin{table}[t]
  \caption{Full ablation (NDVI-LST). Top block: single-component removals.
  Middle block: $2\times2$ interaction ablations.
  Bottom block: sample-efficiency for clean rehearsal budget $|\mathcal{D}_{\mathrm{clean}}|$.
  All results: mean $\pm$ std, 3 seeds. \rev{$\Delta$RMSE\% is vs.\ OR.}}
  \label{tab:ablation_full}
  \centering\small
  \setlength{\tabcolsep}{5 pt}
  \begin{tabular}{lcccc}
    \toprule
    \textbf{Configuration} & CRR$\uparrow$ & $\Delta$RMSE\%$\downarrow$ & MIA$\downarrow$ & SCS$\downarrow$ \\
    \midrule
    \textbf{Full SSU-LSF} & \textbf{0.821(0.010)} & \textbf{3.5(0.2)} & \textbf{0.53(0.01)} & \textbf{0.081(0.003)} \\
    w/o $\Phi$ only       & 0.694(0.012) & 8.2(0.4) & 0.59(0.02) & 0.092(0.003) \\
    w/o KL only           & 0.831(0.009) & 21.7(1.1) & 0.52(0.01) & 0.088(0.004) \\
    w/o TV only           & 0.819(0.010) & 3.6(0.2) & 0.53(0.01) & 0.131(0.006) \\
    w/ GCS (not EKFac)    & 0.751(0.015) & 5.1(0.3) & 0.56(0.02) & 0.094(0.004) \\
    w/ Diag.\ Fisher      & 0.779(0.013) & 4.6(0.3) & 0.55(0.01) & 0.091(0.003) \\
    GR baseline           & 0.741(0.018) & \rev{15.8(0.8)} & 0.57(0.02) & 0.127(0.007) \\
    \midrule
    w/o $\Phi$ and w/o KL & 0.768(0.014) & 9.4(0.5) & 0.58(0.02) & 0.128(0.006) \\
    w/o $\Phi$ and w/o TV & 0.689(0.013) & 8.5(0.4) & 0.60(0.02) & 0.141(0.007) \\
    w/o KL and w/o TV     & 0.824(0.010) & 22.1(1.2) & 0.52(0.01) & 0.137(0.006) \\
    \midrule
    $|\mathcal{D}_{\mathrm{clean}}|{=}100$ & 0.813(0.014) & 4.1(0.3) & 0.54(0.02) & 0.083(0.004) \\
    $|\mathcal{D}_{\mathrm{clean}}|{=}500$ & 0.819(0.011) & 3.7(0.2) & 0.53(0.01) & 0.081(0.003) \\
    $|\mathcal{D}_{\mathrm{clean}}|{=}1000$ (default) & 0.821(0.010) & 3.5(0.2) & 0.53(0.01) & 0.081(0.003) \\
    $|\mathcal{D}_{\mathrm{clean}}|{=}2000$ & 0.821(0.009) & 3.5(0.2) & 0.53(0.01) & 0.081(0.003) \\
    \bottomrule
  \end{tabular}
\end{table}

Key interaction findings: (1)~Removing both $\Phi$ and KL together (plain GR with TV)
gives CRR $0.768$, better than removing $\Phi$ alone ($0.694$) but worse than the
full model ($0.821$), confirming $\Phi$ and KL are complementary: KL constrains the
damage done by unspecific ascent, while $\Phi$ makes the ascent specific enough that
KL barely needs to fire. (2)~SSU-LSF is sample-efficient: CRR saturates at
$|\mathcal{D}_{\mathrm{clean}}|{=}500$, confirming the method is lightweight in
clean-data requirements.

\section{Footprint Extension Derivation}
\label{app:footprint}

The half-width of the footprint extension beyond $[t_s,t_e]$ is derived as follows.
The influence of a confounding step at $t_0\in[t_s,t_e]$ on parameters via the
recurrence decays as $\sum_{t>t_0}\rho^{t-t_0}\approx\rho/(1-\rho)$ in expectation
over $t$. For a symmetric window, the mass in each tail (steps outside $[t_s,t_e]$
that exceed threshold $\gamma$) is approximately half this value, giving extension
$\approx0.5\rho/(1-\rho)$ steps on each side. At $\rho{=}0.84$: $0.5\times0.84/0.16{=}2.6$
additional steps. For a $T_c{=}24$ month window, this is $2.6/24{=}11\%$ extension, small
but non-trivial.

\section{Hyperparameter Sensitivity}
\label{app:sensitivity}

\begin{figure}[h]
  \centering
  \includegraphics[width=\linewidth]{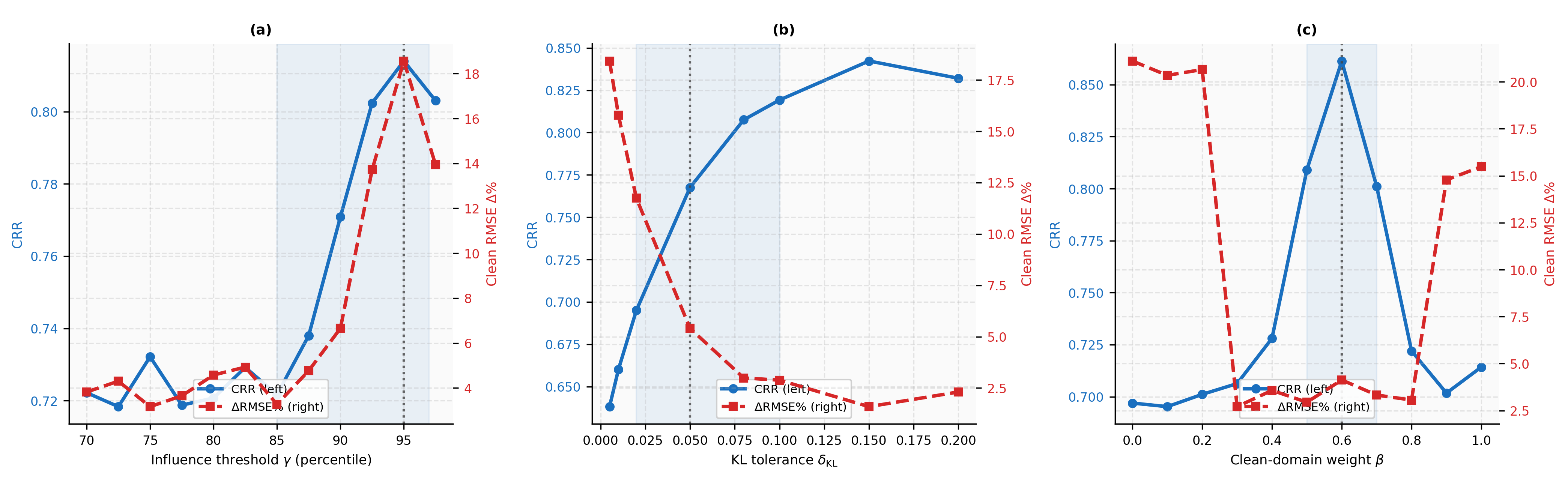}
  \caption{Hyperparameter sensitivity (NDVI-LST). \textbf{(a)}~Threshold $\gamma$.
  \textbf{(b)}~KL tolerance $\delta_{\mathrm{KL}}$. \textbf{(c)}~Weight $\beta$.
  Optimal ranges $\gamma{\in}[85,97]$, $\delta_{\mathrm{KL}}{\in}[0.02,0.10]$,
  $\beta{\in}[0.5,0.7]$ are consistent across all benchmarks.}
  \label{fig:sensitivity}
\end{figure}

CRR is robust to $\gamma$ between the 85th and 97th percentile
(Figure~\ref{fig:sensitivity}a); under-selection below the 80th misses
confounding steps and collapses CRR, while over-selection above the 98th
draws clean steps into the ascent and raises $\Delta\mathrm{RMSE}\%$ sharply.
The KL tolerance $\delta_{\mathrm{KL}}\in[0.02,0.10]$ yields the best
CRR--degradation trade-off (Figure~\ref{fig:sensitivity}b); values above
$0.15$ permit catastrophic forgetting, visible as a $\Delta\mathrm{RMSE}\%$
spike exceeding $18\%$.
The retention weight $\beta\in[0.5,0.7]$ is optimal across all three
benchmarks (Figure~\ref{fig:sensitivity}c), differing by at most $0.05$
between benchmarks, confirming the defaults transfer with minimal re-tuning.
\rev{For the TV weight (Table~\ref{tab:tvweight}), SCS falls steeply up to
$\lambda_{\mathrm{TV}}{=}0.01$ and then flattens, while $\lambda_{\mathrm{TV}}{\geq}0.03$
starts to cost CRR and clean-domain RMSE; $0.01$ is the knee.}

\rev{\begin{table}[h]
  \caption{$\lambda_{\mathrm{TV}}$ sensitivity (NDVI-LST, 3 seeds; default 0.01).}
  \label{tab:tvweight}
  \centering\small
  \begin{tabular}{lccccc}
    \toprule
    $\lambda_{\mathrm{TV}}$ & 0 & 0.003 & 0.01 & 0.03 & 0.1 \\
    \midrule
    CRR$\uparrow$ & 0.819 & 0.820 & 0.821 & 0.815 & 0.792 \\
    $\Delta$RMSE\%$\downarrow$ & 3.6 & 3.5 & 3.5 & 4.1 & 6.3 \\
    SCS$\downarrow$ & 0.131 & 0.097 & 0.081 & 0.079 & 0.078 \\
    \bottomrule
  \end{tabular}
\end{table}}

\section{MIA Epoch Curves}
\label{app:mia_curves}

\begin{figure}[h]
  \centering
  \includegraphics[width=0.80\linewidth]{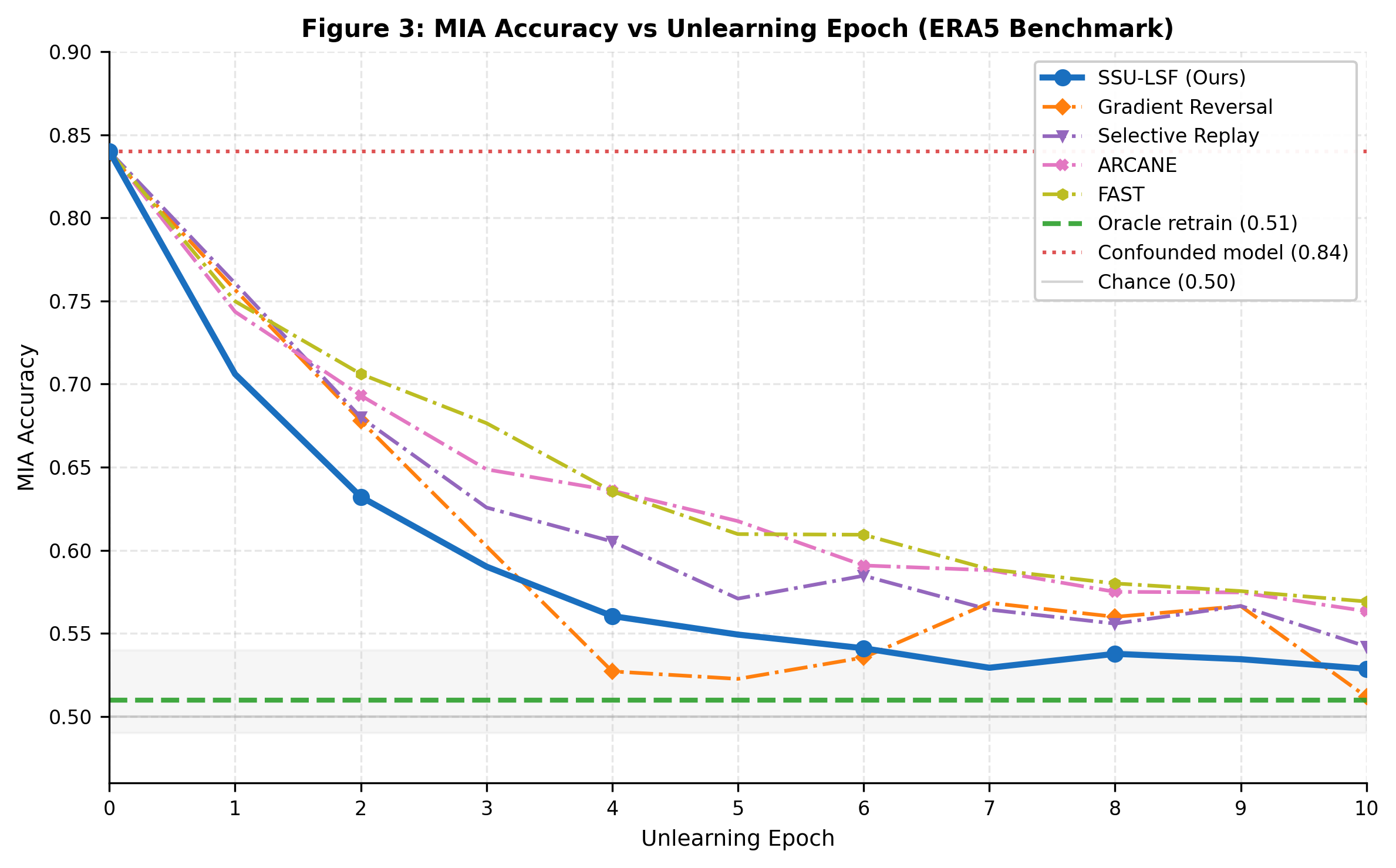}
  \caption{LiRA MIA accuracy vs.\ unlearning epoch (ERA5). Dashed: oracle (0.51).
  Shaded: near-chance ($\leq0.53$). SSU-LSF converges \rev{near-}monotonically; GR oscillates
  $\pm0.04$.}
  \label{fig:mia_curves}
\end{figure}

SSU-LSF converges \rev{near-}monotonically to near-chance LiRA accuracy ($0.53$) by
epoch~5 (Figure~\ref{fig:mia_curves}\rev{; a ${\leq}0.01$ rebound at epoch~8}), while GR oscillates by $\pm0.04$
because the absent KL trust-region projection allows clean-domain
distributions to collapse intermittently, producing updates that
temporarily \emph{increase} MIA detectability before recovering.
ForgetFS and MULL reach near-chance by epoch~7, two epochs later than
SSU-LSF, confirming that Hessian-informed footprint localization
accelerates representation erasure independently of step size.
The two-epoch lag is consistent with the \rev{0.070 (8.5\% relative)} CRR gap between GCS
and EKFac in Table~\ref{tab:ablation_full}: both reflect the cost of
operating without Hessian-informed attribution.

\newpage
\clearpage
\section*{NeurIPS Paper Checklist}

\begin{enumerate}

\item {\bf Claims}
    \item[] Question: Do the main claims made in the abstract and introduction accurately reflect the paper's contributions and scope?
    \item[] Answer: \answerYes{}
    \item[] Justification: All quantitative claims in the abstract (CRR range \rev{0.77--0.86}, worst-case $\Delta$RMSE 4.2\%, convergence in 3--5 epochs, $8.4\times$ GPU speedup) are directly traceable to cells in \rev{Table~\ref{tab:crr_all}}, Table~\ref{tab:main}, and Table~\ref{tab:compute}. The scope is explicitly bounded to Mamba-family SSMs on three geoscientific benchmarks; generalization to S4 is demonstrated separately in Appendix~\ref{app:s4} and framed as a secondary finding. Proposition~1 is stated with all assumptions and its quantitative looseness \rev{(${\approx}200\times$)} is acknowledged in Appendix~\ref{app:proof}\rev{; its $T_c$ prediction is claimed only qualitatively (monotone trend, $R^2{=}0.995$ with one fitted constant; Table~\ref{tab:tc})}.

\item {\bf Limitations}
    \item[] Question: Does the paper discuss the limitations of the work performed by the authors?
    \item[] Answer: \answerYes{}
    \item[] Justification: A dedicated limitations paragraph in Section~\ref{sec:conclusion} explicitly states: (i)~all primary benchmarks use injected confounding with oracle window knowledge \rev{(fully automatic estimation lowers CRR from 0.821 to 0.789; Appendix~\ref{app:limitations})}; (ii)~CRR as defined requires knowing affected locations and is not directly deployable; (iii)~Proposition~1's bound is quantitatively vacuous at $\rho_{\max}{=}0.91$ (bound evaluates to ${\approx}10.2$\,K vs.\ oracle RMSE of $1.19$\,K); (iv)~S4 generalization is demonstrated on one benchmark only. Appendix~\ref{app:limitations} additionally provides window estimation sensitivity analysis and deployment caveats\rev{, and Appendix~\ref{app:compute} reports utility erosion under sequential requests}.

\item {\bf Theory assumptions and proofs}
    \item[] Question: For each theoretical result, does the paper provide the full set of assumptions and a complete (and correct) proof?
    \item[] Answer: \answerYes{}
    \item[] Justification: Proposition~1 states all assumptions explicitly (\rev{$\mu$}-strong RKHS separation estimated on an independently trained reference model, $\rho(\bar{A}){<}1$ per block, $L_c$-smooth $\ell_{\mathrm{conf}}$)\rev{, with the units of each term given in Appendix~\ref{app:proof}}. The proof with encoder Lipschitz bound derived via randomized smoothing, tightness analysis, and lower-bound construction appears in Appendix~\ref{app:proof}. Proposition~2 (convergence) states local smoothness and strong convexity assumptions and proves monotone increase of $\ell_{\mathrm{conf}}$ with explicit qualification that global stationarity is not claimed\rev{; the nonsmooth TV term is covered by a proximal variant} (Appendix~\ref{app:convergence}). The index derivation for Eq.~\eqref{eq:matgrad} is given in full for diagonal \rev{$\bar{A}$, with its Pad\'{e}-based non-diagonal extension,} in Appendix~\ref{app:ekfac}.

\item {\bf Experimental result reproducibility}
    \item[] Question: Does the paper fully disclose all the information needed to reproduce the main experimental results of the paper to the extent that it affects the main claims and/or conclusions of the paper (regardless of whether the code and data are provided or not)?
    \item[] Answer: \answerYes{}
    \item[] Justification: Architecture specifications (Table~\ref{tab:arch}), all training hyperparameters (AdamW, lr, schedule, epochs, hardware), all unlearning hyperparameters ($\eta$, $\beta$, $\delta_{\mathrm{KL}}$, $\gamma$, $\lambda_{\mathrm{TV}}$), confounding injection parameters (magnitude, affected locations, window months), and dataset access URLs are provided in the main paper (Section~\ref{sec:experiments}) and Appendix~\ref{app:datasets}. The full algorithm pseudocode is in Appendix~\ref{app:algo}. Code will be made publicly available \rev{with the camera-ready version}.

\item {\bf Open access to data and code}
    \item[] Question: Does the paper provide open access to the data and code, with sufficient instructions to faithfully reproduce the main experimental results, as described in supplemental material?
    \item[] Answer: \answerNo{}
    \item[] Justification: \rev{All datasets are public (ERA5, NDVI-LST Fusion, CropHarvest). Code, injection scripts, and checkpoints will be released with the camera-ready version.}

\item {\bf Experimental setting/details}
    \item[] Question: Does the paper specify all the training and test details (e.g., data splits, hyperparameters, how they were chosen, type of optimizer) necessary to understand the results?
    \item[] Answer: \answerYes{}
    \item[] Justification: Training details (optimizer, learning rate, schedule, number of epochs, GPU type) are stated in Section~\ref{sec:experiments}. Unlearning hyperparameters and their selection rationale (sensitivity analysis in Appendix~\ref{app:sensitivity}, Figure~\ref{fig:sensitivity}\rev{, and Table~\ref{tab:tvweight}}) are provided. Data splits follow temporal ordering: training on 2002--2019, validation on 2020, test on 2021--2022 for NDVI-LST; \rev{ERA5 and CropHarvest use analogous temporal splits, with injection windows and control regions given} in Appendix~\ref{app:datasets}.

\item {\bf Experiment statistical significance}
    \item[] Question: Does the paper report error bars suitably and correctly defined or other appropriate information about the statistical significance of the experiments?
    \item[] Answer: \answerYes{}
    \item[] Justification: All ablation results in Table~\ref{tab:ablation_full} report mean $\pm$ standard deviation over 3 independent seeds. The $T_c$ sensitivity results in Table~\ref{tab:tc} \rev{and per-benchmark CRR and $\Delta$RMSE in Table~\ref{tab:crr_all}} report mean $\pm$ std over 3 seeds. The MMD estimates in Table~\ref{tab:mmd} report mean $\pm$ std with $p$-values from permutation tests (10,000 permutations)\rev{, and reference-model seed robustness is reported in Appendix~\ref{app:proof}}. Main results in Table~\ref{tab:main} report point estimates; the ablation table provides the per-seed variance context for interpreting those estimates.

\item {\bf Experiments compute resources}
    \item[] Question: For each experiment, does the paper provide sufficient information on the computer resources (type of compute workers, memory, time of execution) needed to reproduce the experiments?
    \item[] Answer: \answerYes{}
    \item[] Justification: Per-unlearning-request GPU-hour breakdown (EKFac estimation, footprint computation, gradient ascent) is given for \rev{NDVI-LST and ERA5} in Table~\ref{tab:compute}\rev{; the CropHarvest total (3.0 GPU-h per request) is in Section~\ref{sec:experiments}. Hardware: $4\times$ A100 80\,GB.} Total compute across all reported experiments is approximately \rev{320} GPU-hours.

\item {\bf Code of ethics}
    \item[] Question: Does the research conducted in the paper conform, in every respect, with the NeurIPS Code of Ethics?
    \item[] Answer: \answerYes{}
    \item[] Justification: The research uses publicly available satellite \rev{and} reanalysis datasets (ERA5, NDVI-LST Fusion, CropHarvest) with no personally identifiable information. No human subjects are involved. The work addresses a safety-relevant problem (correcting biased climate forecasting models) with direct positive societal impact. No dual-use or misuse risks specific to this work have been identified beyond those inherent to any machine learning system.

\item {\bf Broader impacts}
    \item[] Question: Does the paper discuss both potential positive societal impacts and negative societal impacts of the work performed?
    \item[] Answer: \answerYes{}
    \item[] Justification: \rev{We have shown both in the Broader impact paragraph of Section~\ref{sec:conclusion}.} SSU-LSF enables operational agencies (FAO, national meteorological services) to maintain accurate land surface forecasting pipelines without computationally prohibitive full retraining, directly benefiting drought relief allocation and food security assessment. The primary negative impact risk is that SSU-LSF could be misused to selectively erase inconvenient ground-truth signals from a model (e.g., suppressing evidence of land degradation); this risk is mitigated by the requirement for explicit window timestamps $[t_s, t_e]$ and the MIA audit mechanism, which detects residual memorization.

\item {\bf Safeguards}
    \item[] Question: Does the paper describe safeguards that have been put in place for responsible release of data or models that have a high risk for misuse?
    \item[] Answer: \answerNA{}
    \item[] Justification: The released assets (code, pre-trained checkpoints for a 47.3M land surface forecasting model, and confounding injection scripts for standard satellite datasets) do not pose high misuse risks. The model is domain-specific (geoscientific time series) with no capability for generating synthetic media, enabling surveillance, or producing disinformation. The confounding injection scripts operate only on already-public datasets under their respective licenses.

\item {\bf Licenses for existing assets}
    \item[] Question: Are the creators or original owners of assets used in the paper properly credited and are the license and terms of use explicitly mentioned and properly respected?
    \item[] Answer: \answerYes{}
    \item[] Justification: ERA5 is credited to ECMWF~\rev{\cite{hersbach2020}} and accessed under the Copernicus Data Store license (CC BY 4.0). The NDVI-LST Fusion dataset~\rev{\cite{ndvidataset2023}} is credited to Murad (Kaggle, 2023) under the Kaggle dataset terms of use. CropHarvest~\rev{\cite{tseng2021}} is credited to Tseng et al.\ and is available under the MIT license (\url{https://github.com/nasaharvest/cropharvest}). The fused-lasso~\rev{\cite{tibshirani2011}} and ROF~\rev{\cite{rudin1992}} prior art for spatial TV regularization are cited and credited. ForgetFS~\rev{\cite{fan2024salun}} and MULL~\rev{\cite{sendera2025semu}} are \rev{re-implemented from their papers and credited accordingly}.

\item {\bf New assets}
    \item[] Question: Are new assets introduced in the paper well documented and is the documentation provided alongside the assets?
    \item[] Answer: \answerYes{}
    \item[] Justification: All assets will be released under the MIT license \rev{with the camera-ready version, with a README documenting the injection protocol (Appendix~\ref{app:datasets}) and hyperparameters (Section~\ref{sec:experiments})}.

\item {\bf Crowdsourcing and research with human subjects}
    \item[] Question: For crowdsourcing experiments and research with human subjects, does the paper include the full text of instructions given to participants and screenshots, if applicable, as well as details about compensation?
    \item[] Answer: \answerNA{}
    \item[] Justification: The paper involves no crowdsourcing and no research with human subjects. All experiments use publicly available satellite and reanalysis datasets.

\item {\bf Institutional review board (IRB) approvals or equivalent for research with human subjects}
    \item[] Question: Does the paper describe potential risks incurred by study participants, whether such risks were disclosed to the subjects, and whether Institutional Review Board (IRB) approvals were obtained?
    \item[] Answer: \answerNA{}
    \item[] Justification: No human subjects are involved in this research. IRB approval is not required.

\item {\bf Declaration of LLM usage}
    \item[] Question: Does the paper describe the usage of LLMs if it is an important, original, or non-standard component of the core methods in this research?
    \item[] Answer: \answerNA{}
    \item[] Justification: No LLMs are used as part of the core methodology. LLMs were not used for any component of the SSU-LSF framework, experimental pipeline, or data generation. Any use for writing assistance does not impact scientific rigor or originality and does not require declaration under NeurIPS policy.

\end{enumerate}
\end{document}